\documentclass[conference]{IEEEtran}
\IEEEoverridecommandlockouts
\usepackage{cite}
\usepackage{amsmath,amssymb,amsfonts}
\usepackage{algorithmic}
\usepackage{graphicx}
\usepackage{textcomp}
\usepackage{xcolor}
\usepackage{tabularx}
\usepackage{booktabs} 
\usepackage{subcaption}
\usepackage{hyperref}
\usepackage{textcomp}
\def\BibTeX{{\rm B\kern-.05em{\sc i\kern-.025em b}\kern-.08em
    T\kern-.1667em\lower.7ex\hbox{E}\kern-.125emX}}
\begin{document}

\title{Do CNNs Encode Data Augmentations?
}

\author{\IEEEauthorblockN{Eddie Yan*\thanks{*Work done while an intern at Google Brain. Published as a conference paper at IJCNN 2021.}}
\IEEEauthorblockA{
\textit{Allen School}\\
\textit{University of Washington}\\
Seattle, USA \\
eqy@cs.washington.edu}
\and
\IEEEauthorblockN{Yanping Huang}
\IEEEauthorblockA{\textit{Google Brain}\\
Mountain View, USA \\
huangyp@google.com}
}

\maketitle

\begin{abstract}
Data augmentations are important ingredients in the recipe for training robust neural networks, especially in computer vision.
A fundamental question is whether neural network features encode data augmentation transformations.
To answer this question, we introduce a systematic approach to investigate which layers of neural networks are the most predictive of augmentation transformations.
Our approach uses features in pre-trained vision models with minimal additional processing to predict common properties transformed by augmentation (scale, aspect ratio, hue, saturation, contrast, and brightness).
Surprisingly, neural network features not only predict data augmentation transformations, but they predict many transformations with high accuracy.   
After validating that neural networks encode features corresponding to augmentation transformations, we show that these features are encoded in the early layers of modern CNNs, though the augmentation signal fades in deeper layers.
\end{abstract}


\section{Introduction}
\begin{figure}[t]
\centering
\begin{subfigure}[h]{0.5\textwidth}
\centering
\includegraphics[width=1.0\textwidth]{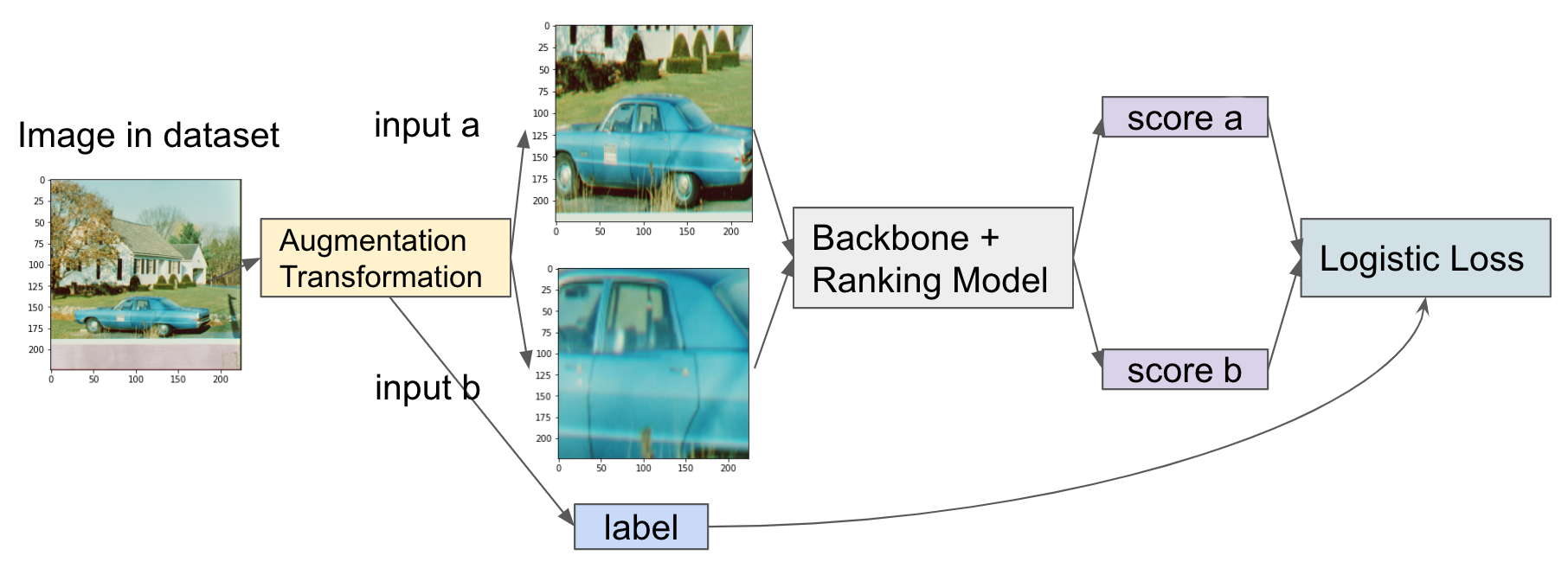}
\caption{Our model pipeline; a ranking model uses backbone features to score (rank) the extent of augmentations in pairs of inputs. Here, each score is a measure of the extent of zoom in each input image.}
\label{fig:overview}
\end{subfigure}
\begin{subfigure}[h]{0.5\textwidth}
\includegraphics[width=\textwidth]{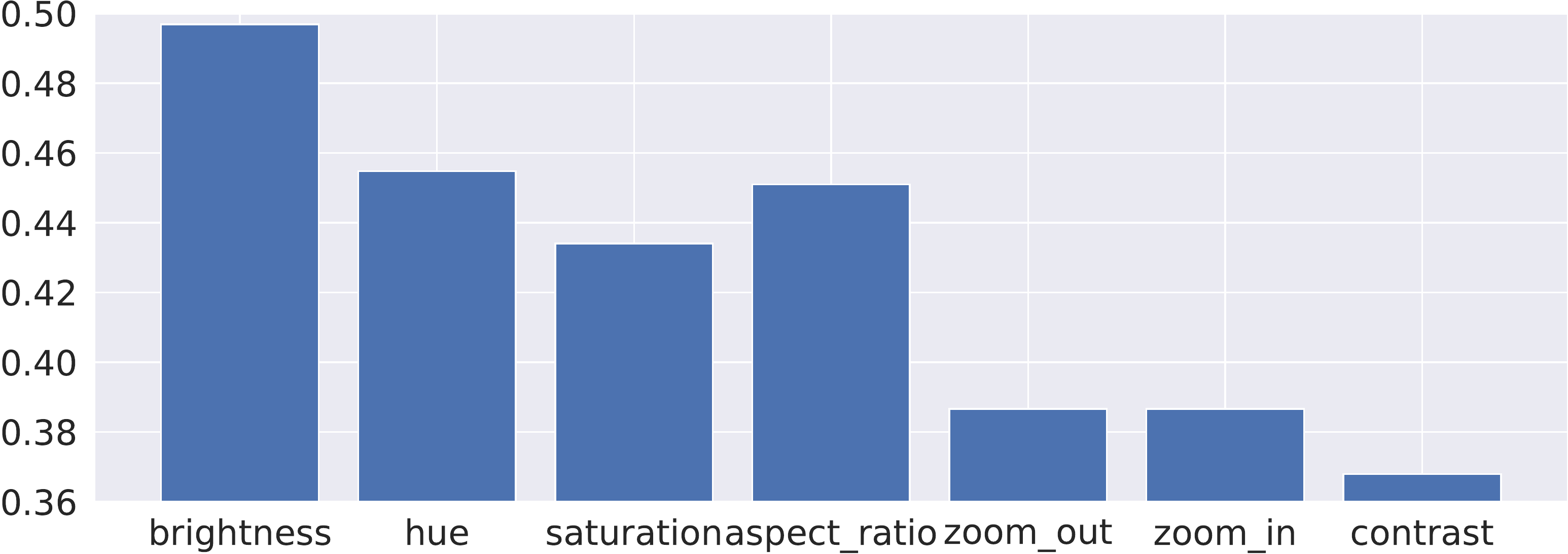}
\caption{Relative importance of the first block of ResNet-18 for predicting each of the data augmentation ranking tasks. ($x$-axis ordered roughly by complexity of augmentations).}
\label{fig:importance}
\end{subfigure}
\caption{Do layer activations from CNNs encode input variations introduced by data augmentation? For a given image, a pair of inputs is generated by varying the extent of a data augmentation (e.g., scale). ~
\autoref{fig:overview} shows how we probe model features for augmentation information. ~\autoref{fig:importance} shows that early ResNet layers are more important for encoding low-level transformations (brightness and saturation).}
\label{fig:training}
\end{figure}
Convolutional neural networks (CNNs) have enjoyed tremendous success on popular computer vision problems.
Ideally, vision models for these tasks would be equivariant to perturbations in color, translation, scale, and rotation.
Translation invariance has been partially architected in CNNs~\cite{pmlr-v97-zhang19a}, and building models with other equivariant properties is an active area of research~\cite{xu2014scale, kanazawa2014locally, cohen2016group, sosnovik2019scale, zhu2019scale, kondor2018generalization}.
In spite of their success, CNN models remain sensitive to small changes~\cite{goodfellow2014explaining} in training data with respect to desirable equivariants.
The typical~\cite{krizhevsky2012imagenet}, yet effective ~\cite{zhang2016understanding} approach to build robust models is to leverage brute force via data augmentation.

However, current understanding of the effects of data augmentations is limited, and using data augmentations often requires ad-hoc or task-specific heuristics.
An instance of this problem occurs when objects are shown to models at different scales: popular models for classification exhibit a noticeable drop in accuracy when the scale of their test-time data does not match that of their training-time data~\cite{NIPS2019_9035}.
In parallel, we observe that enhanced data augmentation can lead to dramatic improvements in accuracy, especially in adversarial scenarios~\cite{xie2019adversarial}, but this requires rearchitecting models to effectively leverage adversarial examples.

The importance of data augmentation leads to natural questions about what useful concepts models learn from data augmentations.
As data augmentations are often intended to reflect natural priors (e.g., objects belonging to the same class have variations in scale), a relevant question is how these priors are captured by the model.
Concretely, we ask whether variations corresponding to data augmentations are encoded by models, and where this encoding takes place.
For example, do models encode brightness variations in the earlier layers, in the later layers, or both? 
Which data augmentations correspond to \emph{low-level} model features, and which correspond to \emph{high-level} model features?


We search for answers to these questions by investigating whether intermediate activations of models capture input differences introduced by data augmentation.
First, we define a set of attributes (scale, aspect ratio, and color transformations) that are desirable invariants (equivariants) for models and commonly targeted by the data augmentations of current model training pipelines~\cite{cubuk2019autoaugment}.
Following these definitions, 
we propose several experiments, introducing a data augmentation ranking task (as illustrated in Figure~\ref{fig:overview}) to understand whether CNNs implicitly learn a representation for these attributes, comparing against baseline models relying on primitive features.
These experiments measure the predictive performance of a ranking model that uses intermediate features collected from pre-trained models to predict augmentation attributes. Following these experiments, we inspect the relative importance of features used in the ranking model to understand the relative importance of layers in encoding data augmentation attributes.

Our results show that CNNs implicitly learn to encode attributes of popular data augmentations such as scale, aspect ratio, saturation, and contrast without being explicitly trained on these objectives.
Additionally, we find that these attributes are typically encoded in the earlier layers of networks, suggesting that models learn to normalize input variations introduced by data augmentations. 
Later layers appear more important for aspect ratio and scale, which can be considered higher-level than attributes such as brightness and saturation, as shown in Figure~\ref{fig:importance}.
We present data augmentation prediction as tool to improve the currently limited interpretability~\cite{lipton2018mythos} of CNNs.

\section{Related Work}
Data augmentations are a tried and true method of improving CNN accuracy~\cite{ciregan2012multi, krizhevsky2012imagenet}.
Prior work has also compared data augmentation in the input space with augmentations applied in the feature space of neural networks, with the conclusion that ``plausible transformations'' that are guaranteed to avoid changing the label yield the most improvement in model performance~\cite{wong2016understanding}. 
More recently, using augmentations to incrementally increase the difficulty of training~\cite{2019arXiv191104252X}, automatically generating augmentation strategies~\cite{cubuk2019autoaugment}, and modifying networks to better support adversarial or corruption-based augmentations~\cite{xie2019adversarial} have emerged as promising directions.
Augmentations are relevant in the semi-supervised setting~\cite{berthelot2019mixmatch}, label-smoothing~\cite{zhang2017mixup}, regularization~\cite{yun2019cutmix, devries2017improved}, and dataset watermarking~\cite{alex2020radioactive}.
Work has also been done to understand the theoretical motivation behind data augmentation~\cite{dao2019kernel, chen2019grouptheoretic}.
Data augmentations are also important from the perspective of model biases such as texture and shape bias~\cite{hermann2019origins}.

On the side of neural network understanding, visualizing features and saliency maps~\cite{erhan2009visualizing, simonyan2013deep, zeiler2014visualizing, Zhou_2016_CVPR, selvaraju2017grad} have enabled interpretation of the functionality and learned patterns of neural network layers.
Intermediate model features have also been used to synthesize and visualize the textures learned by models ~\cite{lin2016visualizing, gatys2015texture}.
Automated approaches such as training classifiers to infer brain activity and state are a longstanding staple of neuroscience research~\cite{pereira2009machine}, and have been co-opted recently for understanding fundamental questions about what is encoded in neural network activations~\cite{islam2019much}.

The challenges of choosing the best model architecture for a task and scaling it appropriately~\cite{pmlr-v97-tan19a} have emerged as important problems, yet both model architecture and model capacity are usually treated as black-box parameters~\cite{zoph2016neural, real2019regularized}.
By investigating how different components of models react to data augmentation, we hope to reveal which components of models are relevant for good classification performance.

\section{A Ranking Model for Augmentations}
To assess whether neural network features encode data augmentation transformations, we propose a ranking task that predicts the \textit{relative} extent of augmentation attributes given intermediate neural network features.
We employ a ranking model instead of a regression approach since obtaining the \textit{absolute} extent of augmentation is difficult.
For example, for the task of predicting the scale of an object, it is difficult to design a numerical definition of scale that is consistent across many different input examples and object classes.
We use a separate ranking model as it facilitates interpretability over blackbox approaches that only consider the final output or accuracy of model predictions.
As we show in~\autoref{sec:importance}, we can leverage the ranking model weights to infer the importance of different layers to the ranking tasks.

\label{sec:ranking_tasks}
To circumvent the requirement of precisely-labeled data for augmentation attributes, we only attempt to rank the relative values of augmentation attributes using pairwise rank-loss~\cite{chen2009ranking}, which can be considered a binary classification task for pairs of input examples.
For the case of scale, the task is to decide whether the scale of the object in one image is greater than the scale of the object in the other.
More formally, for each $i,j$ pair of examples the loss function is defined as
$$\log{(1 + \exp(-\text{sgn}(v_i - v_j) \times (f(x_i)-f(x_j))}))$$
where $v_i, v_j$, $x_i, x_j$, and $f$ denote the true augmentation parameters, inputs to the ranking model, and ranking model respectively.
For each image in the dataset, we produce ordered pairs of images by applying an augmentation transformation parameterized by different random values.


\section{Choosing and Defining Augmentations}
\label{sec:augmentations}
\begin{figure}
\begin{tabular}[c]{ccc}
    \begin{subfigure}[h]{0.12\textwidth}
    \centering
    \includegraphics[width=\textwidth]{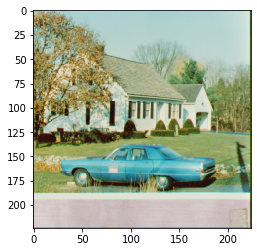}
    \end{subfigure}&
    \begin{subfigure}[h]{0.12\textwidth}
    \centering
    \includegraphics[width=\textwidth]{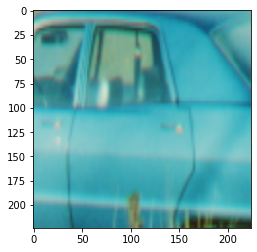}
    \end{subfigure}&
    \begin{subfigure}[h]{0.12\textwidth}
    \centering
    \includegraphics[width=\textwidth]{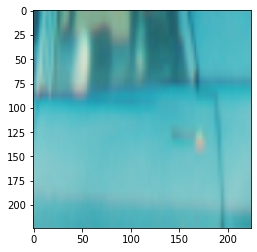}
    \end{subfigure}\\
    \begin{subfigure}[h]{0.12\textwidth}
    \centering
    \includegraphics[width=\textwidth]{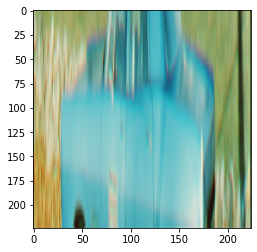}
    \end{subfigure}&
    \begin{subfigure}[h]{0.12\textwidth}
    \centering
    \includegraphics[width=\textwidth]{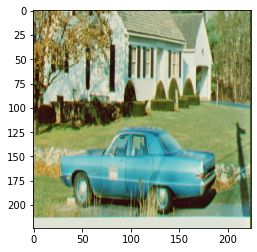}
    \end{subfigure}&
    \begin{subfigure}[h]{0.12\textwidth}
    \centering
    \includegraphics[width=\textwidth]{visualizations/img1.png}
    \end{subfigure}\\
    \begin{subfigure}[h]{0.12\textwidth}
    \centering
    \includegraphics[width=\textwidth]{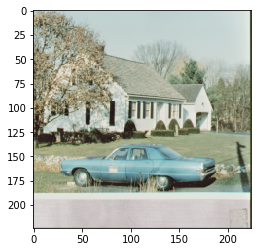}
    \end{subfigure}&
    \begin{subfigure}[h]{0.12\textwidth}
    \centering
    \includegraphics[width=\textwidth]{visualizations/img1.png}
    \end{subfigure}&
    \begin{subfigure}[h]{0.12\textwidth}
    \centering
    \includegraphics[width=\textwidth]{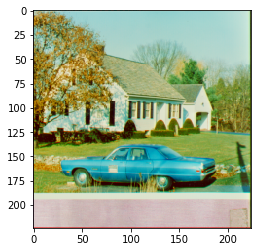}
    \end{subfigure}
\end{tabular}
    \caption{Examples of our definition of scale (row 1), aspect ratio (row 2), and saturation (row 3). We order columns by the extent of the augmentation transformation in each example from left to right.
    \label{fig:example}}
\end{figure}
We describe our definitions of scale, aspect ratio, hue, contrast, saturation, and brightness in this section, focusing on the constraint that our definitions must yield an ordering or ranking of input examples. ~\autoref{fig:example} shows examples for some augmentations considered.
We choose these augmentations based on the following criteria:
(1) Ease of implementation: given an unlabeled set of images, it is straightforward to infer an ordering of perturbed images?
(2) Popularity in training pipelines: each of the transformations considered are either partially or fully implemented in standard TensorFlow~\cite{abadi2016tensorflow}.
(3) Diversity in abstraction level: scale and aspect ratio are higher level attributes, whereas color attributes can almost be directly inferred from raw pixel values with limited context.

\subsection{Scale}
\label{sec:scale}
We carefully settle on a narrow definition of object scale, avoiding semantic definitions of scale, especially between different object classes.
For example, we are not attempting to assess whether models capture facts such as ``elephants are bigger than dogs.''
We choose a pragmatic definition of scale corresponding more closely to the \emph{solid angle} of an object or the proportion of the field of view occupied by an object.

This definition of scale captures the issue of the ``train-test resolution discrepancy''~\cite{NIPS2019_9035}, where test-time crops of images that occupy a smaller area than training-time crops reduce model accuracy and is reflected in the random cropping augmentation that is commonly used to present objects of different scales at training time.
This definition is distinct from resolution or fidelity; one can craft arbitrary examples where both high and low resolution versions of the same image map to the same scale after they are cropped and resized.

Additionally, we add the qualification that we consider scale to be invariant to occlusion or cropping as long as the object is still partially visible in the frame.
We use this qualification to disentangle scale from the related but separate concept of \emph{bounding-box area} occupied by an object.
~\autoref{fig:example} gives examples following our definition of scale. 

From this definition of scale, we define two ranking tasks: ``zoom-out'' and ``zoom-in.''
For the ``zoom-out'' task, we generate pairs of input images that zoom-out from the bounding boxes of objects to generate input images with different scales.
We uniformly sample two values in the range $[0.1, s]$, where $s$ is the smallest of the total vertical or horizontal distance from the border of the bounding box to the boundaries of the image.
For the images in the dataset (\autoref{sec:dataset}) we use, $s$ is expected to be at least $0.3$.
For the ``zoom-in'' task, the different scales are generated by zooming-in on bounding boxes to different extents.
We uniformly sample two values in $[0.5, 0.9]$ that determine the fraction of the bounding box to trim before resizing the result to the input size of the backbone model ($224\times224$) for each pair of inputs.
We define the zoom-in and zoom-out tasks separately because although they may be of similar difficulty for a human evaluator, intuitively the zoom-out task may be easier as the area occupied by an object is a proxy for scale when the object of interest does not occupy the entire frame.

\subsection{Aspect Ratio}
Models are naturally exposed to a range of object aspect ratios at training time through random cropping and natural variation in the input distribution.
Random cropping is an important source of aspect ratio variation, as many augmentation pipelines do not consider the original aspect ratios of objects as constraints on the crop dimensions.
With respect to aspect ratio, we define the ranking order from wide to thin, or the ratio of vertical to horizontal pixels present in the input after cropping (but before resizing).
Note that while ordering the aspect ratio between two arbitrary objects is difficult, this definition suffices when only considering different crops of the same object.

The aspect ratio task uses the same pipeline as the scale tasks, with the objective changed to ranking the ratio of vertical to horizontal pixels.
To generate each input image, we sample four random uniform values in $[0.4, 0.4]$ that determine the proportion of horizontal and vertical pixels to trim.

\subsection{Hue, Saturation, Contrast, Brightness}
\label{sec:define_color}
Hue, saturation, and contrast are common distortions applied to input images.
As each of these augmentations are parameterized by either relative multipliers or absolute deltas to the original image, these parameters lend themselves naturally to an ordering for ranking.
We include brightness as a sanity check that should be trivially encoded for both the CNN backbones and baselines.
While we consider contrast a color transformation, it is arguably higher-level than the other augmentations as discerning contrast requires context.

As before, we sample random uniform values for each of these tasks.
For saturation and contrast, we sample the relative multipliers used to apply the transformation to determine the ranking labels (in the range $[0.5, 1.5]$).
For hue, we rank the delta relative to the original image (in the range $[-0.2, 0.2]$).


\begin{figure}
    \centering
    \begin{subfigure}[b]{0.12\textwidth}
    \centering
    \includegraphics[width=\textwidth]{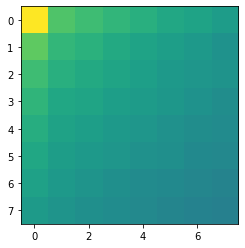}
    \caption{}
    \label{fig:my_label}
    \end{subfigure}
    \begin{subfigure}[b]{0.12\textwidth}
    \centering
    \includegraphics[width=\textwidth]{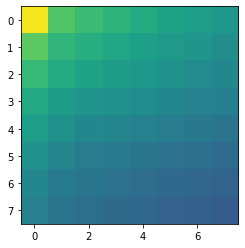}
    \caption{}
    \label{fig:my_label}
    \end{subfigure}
    \begin{subfigure}[b]{0.12\textwidth}
    \centering
    \includegraphics[width=\textwidth]{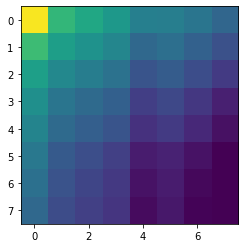}
    \caption{}
    \end{subfigure}
    \caption{Average magnitude (brighter indicates higher magnitude) of frequency coefficients from an $8 \times 8$ patch-wise DCT on images at increasing scales (left to right). Coefficients are ordered by frequency in a zig-zag pattern, with lowest in the top left and highest in the bottom right. The magnitude of higher frequency coefficients decreases as scale increases.}
    \label{fig:dct}
\end{figure}

\section{Methodology}
To understand whether CNN activations capture attributes of data augmentations, we adopt an experiment pipeline similar to one used to extract position information from CNNs~\cite{islam2019much};
we also use the intermediate activations as input to a predictor from a pre-trained vision model with frozen parameters, but with several key differences.
Instead of attempting to generate a two-dimensional output, our prediction task is learning to rank input examples according to their data augmentations.
Our ranking model uses only average pooling and a single linear layer to allow for easy interpretation of the model weights.
For position information, the ground-truth can be generated deterministically, and it is the same across all images.
However, in the case of data augmentations, ranking labels are generated on-the-fly, in tandem with the augmentations.



\subsection{Dataset}
\label{sec:dataset}
We use a subset of the ImageNet~\cite{deng2009imagenet} training dataset in our experiments.
We limit our subset to images that have exactly one bounding box to mitigate the effect of cropping only some relevant objects in view.
We also choose images with bounding boxes that span at least $30\%$ of the input image, with the additional requirement that the borders of the bounding box must be at least $30\%$ of the image dimensions away from edges of the image.
Together, these requirements ensure that there is range to zoom out from bounding boxes and to provide reasonable resolution when zooming in on a bounding box.
These constraints reduce the original $1.2$ million ImageNet images to roughly $86,000$ images, which we split into $65,000$ and $21,000$ images for training and validation.
For simplicity, we use this dataset for all of our ranking tasks, even those that do not require bounding box constraints.

\subsection{Baseline Comparisons}
We also evaluate two baselines that either use an $8\times8$ discrete cosine transform (DCT) to generate features (to understand the impact of frequency information), or are passed the input images directly (passthrough).
~\autoref{fig:dct} shows an example of how the magnitudes of frequency coefficients change with the scale of an object.
For the DCT baseline, we apply average pooling to the DCT features while the spatial dimensions of the passthrough baseline are not reduced.

\subsection{Ranking Model and Training Pipeline}
\begin{figure}[t]
\includegraphics[width=0.40\textwidth]{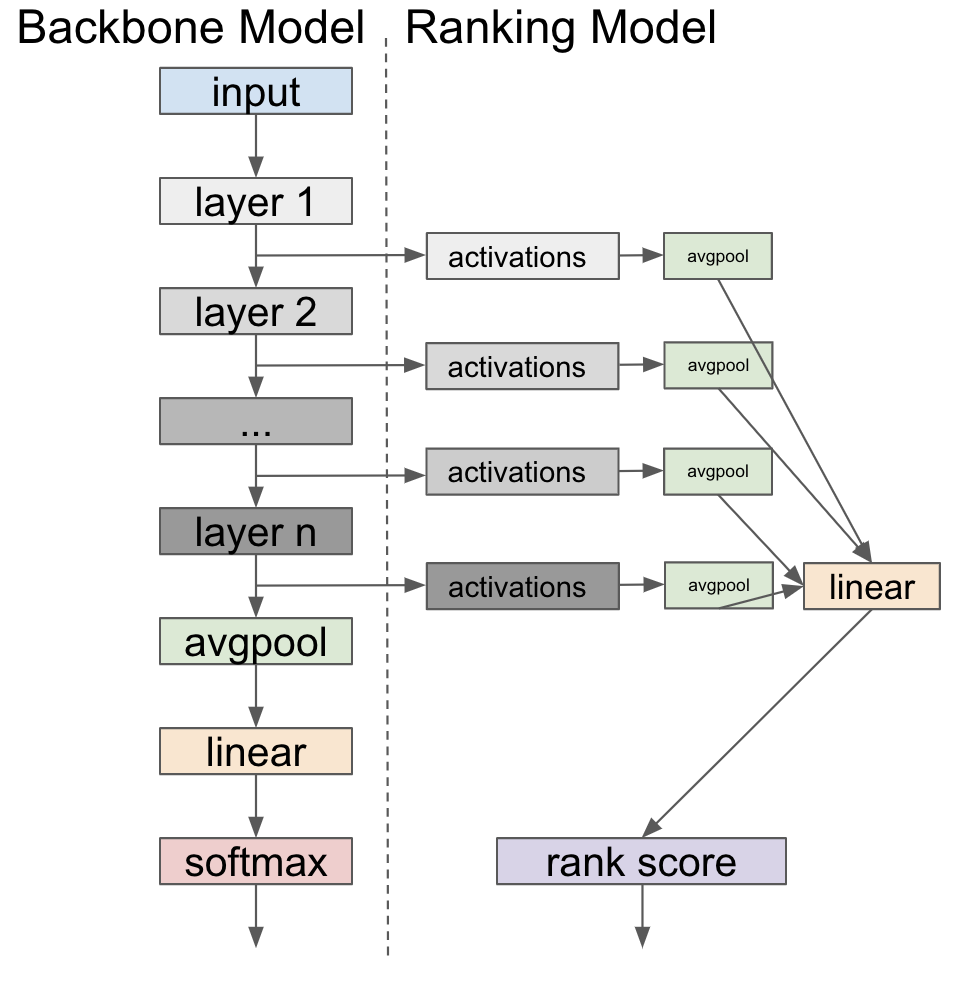}
\caption{Backbone and ranking model used in our evaluation. The backbone is a pre-trained CNN (such as ResNet-18), with parameters frozen. The activations from the backbone are average-pooled to align their spatial dimensions and fed to a linear layer that produces a score that ranks input pairs (e.g., score a $>$ score b?).}
\label{fig:ranking}
\end{figure}

Our ranking model uses the intermediate activations from a pre-trained CNN as inputs to rank instances of a given data augmentation transformation.
~\autoref{fig:ranking} shows a high-level diagram of the relationship between the backbone model and the ranking model.
For our experiments, we use ResNet-18/50~\cite{he2016deep} as the backbone, although this approach is compatible with any feedforward CNN.
To unify the spatial dimensions of each layer, we apply global average pooling while preserving the channels.
The average-pooled tensors are then fed to a single linear layer that computes the ranking score for a given input example.
For each each pair of input examples, we use the ranking scores and logistic loss to fit the linear layer.

The training pipeline begins with iteration through a dataset of images, where each image is used to generate a pair of input examples.
Each input example is transformed by sampling a random variable and the current augmentation ranking task (e.g., scale).
At this time, a label for this pair of input examples can be computed as a boolean expression of the random variables (e.g., scale\_a $>$ scale\_b?).
A collection of pairs and labels comprise a batch that is used to fit the linear layer with logistic loss.
The parameters of the backbone model are frozen during training of the ranking model to prevent the ranking task from affecting the intermediate features of the backbone.
We use the same approach with the baselines, with average pooling omitted for the passthrough baseline.

\subsection{Where are data augmentations encoded?}
We use the weights of the linear layer to measure the relative importance of the activations for each layer of the backbone model.
Due to the simplicity of the linear ranking model, we can measure the contribution of each layer of the backbone by taking the product of the weights and the corresponding standard deviations in the layer activations.

\section{Evaluation}
We begin the evaluation with the accuracy results (\autoref{tab:accs}) for each of the pairwise ranking tasks.
Due to the binary nature of a pairwise ranking task, the accuracy of random guessing is $50\%$.
For all tasks, we find that the ResNet backbones either match or substantially outperform the baselines, particularly on the augmentations that are not color manipulations. This suggests CNNs may implicitly model scale and aspect ratio as components of features.

Prior work has compared the early layers of CNN to the discrete cosine transform (DCT) ~\cite{gueguen2018faster}.
To some extent, we expect the DCT~(\autoref{fig:dct}) and low-level features of earlier layers to act as a proxy for scale and/or aspect ratio.
Intuitively, two views of the same object at different scales are expected to contain different frequency domain representations, where the smaller scale view is expected to have more high frequency components than the larger scale view.
If CNNs capture some elements of frequency domain transforms in convolution layers, we would expect that this information could be used to better infer scale information.
Other augmentations, such as hue and saturation, may present cues in the absolute or relative values of the color channels early in network architectures.

\begin{table}[t]
\begin{footnotesize}
\begin{center}
\begin{tabularx}{0.5\textwidth}{  | >{\raggedright\arraybackslash}X 
  | >{\centering\arraybackslash}X 
  | >{\raggedleft\arraybackslash}X |} 
 \hline
  & Zoom In-Train & Zoom In-Val \\ 
 \hline
 Passthrough & 97.7 & 46.4 \\ 
 \hline
 DCT & 56.8 & 46.6 \\ 
 \hline
 ResNet-18 & 93.9 & \textbf{90.1} \\ 
 \hline
 ResNet-50 & 90.8 & 84.9 \\ 
 \hline
\end{tabularx}
\begin{tabularx}{0.5\textwidth}{  | >{\raggedright\arraybackslash}X 
  | >{\centering\arraybackslash}X 
  | >{\raggedleft\arraybackslash}X |} 
 \hline
  & Zoom Out-Train & Zoom Out-Val\\ 
 \hline
 Passthrough & 98.5 & 51.8 \\ 
 \hline
 DCT & 57.5 & 52.4 \\ 
 \hline
 ResNet-18 & 82.4 & \textbf{68.8} \\ 
 \hline
 ResNet-50 & 77.6 & 64.8 \\ 
 \hline
\end{tabularx}
\begin{tabularx}{0.5\textwidth}{  | >{\raggedright\arraybackslash}X 
  | >{\centering\arraybackslash}X 
  | >{\raggedleft\arraybackslash}X |} 
 \hline
  & Aspect Ratio-Train & Aspect Ratio-Val\\ 
 \hline
 Passthrough & 98.7 & 54.9 \\ 
 \hline
 DCT & 54.1 & 57.7 \\ 
 \hline
 ResNet-18 & 87.6 & 80.9 \\ 
 \hline
 ResNet-50 & 85.9 & \textbf{81.3} \\ 
 \hline
\end{tabularx}
\begin{tabularx}{0.5\textwidth}{  | >{\raggedright\arraybackslash}X 
  | >{\centering\arraybackslash}X 
  | >{\raggedleft\arraybackslash}X |} 
 \hline
  & Hue-Train & Hue-Val\\ 
 \hline
 Passthrough & 94.0 & 65.0 \\ 
 \hline
 ResNet-18 & 87.6 & \textbf{71.6} \\ 
 \hline
 ResNet-50 & 84.0 & 66.0 \\ 
 \hline
\end{tabularx}
\begin{tabularx}{0.5\textwidth}{  | >{\raggedright\arraybackslash}X 
  | >{\centering\arraybackslash}X 
  | >{\raggedleft\arraybackslash}X |} 
 \hline
  & Saturation-Train & Saturation-Val\\ 
 \hline
 Passthrough & 97.5 & 98.9 \\ 
 \hline
 ResNet-18 & 97.5 & 98.3 \\ 
 \hline
 ResNet-50 & 95.2 & 94.0 \\ 
 \hline
\end{tabularx}
\begin{tabularx}{0.5\textwidth}{  | >{\raggedright\arraybackslash}X 
  | >{\centering\arraybackslash}X 
  | >{\raggedleft\arraybackslash}X |} 
 \hline
  & Contrast-Train & Contrast-Val\\ 
 \hline
 Passthrough & 100.0 & 62.0 \\ 
 \hline
 ResNet-18 & 100.0 & \textbf{100.0}\\ 
 \hline
 ResNet-50 & 99.7 & 99.7\\ 
 \hline
\end{tabularx}
\begin{tabularx}{0.5\textwidth}{  | >{\raggedright\arraybackslash}X 
  | >{\centering\arraybackslash}X 
  | >{\raggedleft\arraybackslash}X |} 
 \hline
  & Brightness-Train & Brightness-Val\\ 
 \hline
 Passthrough & 100.0 & 100.0 \\ 
 \hline
 ResNet-18 & 100.0 & 100.0 \\ 
 \hline
 ResNet-50 & 99.3 & 98.8 \\ 
 \hline
\end{tabularx}
\end{center}
\end{footnotesize}
\caption{Accuracies for ranking models that use the baselines and ResNet backbones across the ranking tasks. ResNet features encode many augmentation attributes to a high degree of accuracy, particularly high-level ones such as scale and aspect ratio. ResNet features also beat the baselines on contrast by a wide margin. The accuracy of the ranking model can be used as a proxy to determine to what degree an augmentation attribute is encoded in the CNNs.}
\label{tab:accs}

\end{table}
When comparing results for the scale tasks, we note that the performance of the ResNet backbone was substantially lower for the ``zoom-out'' than ``zoom-in'' task.
This drop in accuracy was surprising as it was thought that the ranking model could rely on the later layers and localization as a proxy for scale, although it is possible that the use of average pooling in the ranking model could have limited localization information.
Additionally, performance on the zoom-out task may have suffered as a consequence of it being more fine-grained than the zoom-in task: many images may have a limited amount of slack in which crop sizes can be increased without overstepping image boundaries.
Still, the performance of the ResNet backbones far surpassed the DCT baseline on both scale tasks, suggesting that CNNs have stronger cues for object scale than spatial frequency.

This result suggests another source of scale information may appear in the higher-level representations of networks.
With the knowledge that activations late in CNNs (e.g., at the last layer) map neatly to class labels~\cite{Zhou_2016_CVPR}, it is plausible that high-level features map coarsely to scale as well (e.g., by their spatial extent in the last layer).
However, we attempt to avoid trivial cues for scale via a very simple ranking model (\autoref{fig:ranking}) and by reducing spatial information via pooling.



Across some tasks, we observe that ranking using the ResNet-18 backbone sometimes outperforms the ResNet-50 backbone.
We suspect that this is due to the large increase in the number of input dimensions to the ranking model when ResNet-50 is used (due to the increase in total number of channels), and regularizing the weights of the ranking model could yield improved performance.
The heavy overfitting of the passthrough baseline can likely be attributed to reliance on absolute position (no average pooling is used) that is not generalizable to the validation set.
An alternative hypothesis is that ResNet-50 yields lower ranking performance because it more successfully normalizes away perturbations caused by augmentations.
This hypothesis is interesting as it suggests that models with stronger performance may do a better job of eliminating differences created by data augmentations.

Hue appears to be the least favorable task for the ResNet backbones (relative to the baselines).
We suspect that this may be due to the narrow range of hue considered, or the difficultly in assessing the absolute delta in hue from the original image.
We expect the task of ranking the raw value of hue rather than the magnitude to be easier.
On the opposite end, contrast appears to be the least favorable task for the baselines (relative to the ResNet backbones), especially of the color augmentations.
We expect that this is because contrast requires more image context and consequentially is a higher-level attribute than hue or saturation.
Accordingly, contrast depends more on later layers of the backbones than the other color transformations~(\autoref{fig:resnet18and50}).

We find that the baseline backbones achieve their highest performance on the color tasks.
This is relatively unsurprising, as some color attributes (such as saturation) may be discernible by the raw values of the input color channels.
More surprisingly, however, was that while the early layers were favored especially for the color-focused transformations, the most highly weighted layer was not the stem of the ResNet models but rather a few layers later.

\subsection{Which layers encode the augmentations?}
\label{sec:importance}
\begin{figure*}
\begin{center}
\begin{subfigure}[H]{0.40\textwidth}
    \centering
    \includegraphics[width=\textwidth]{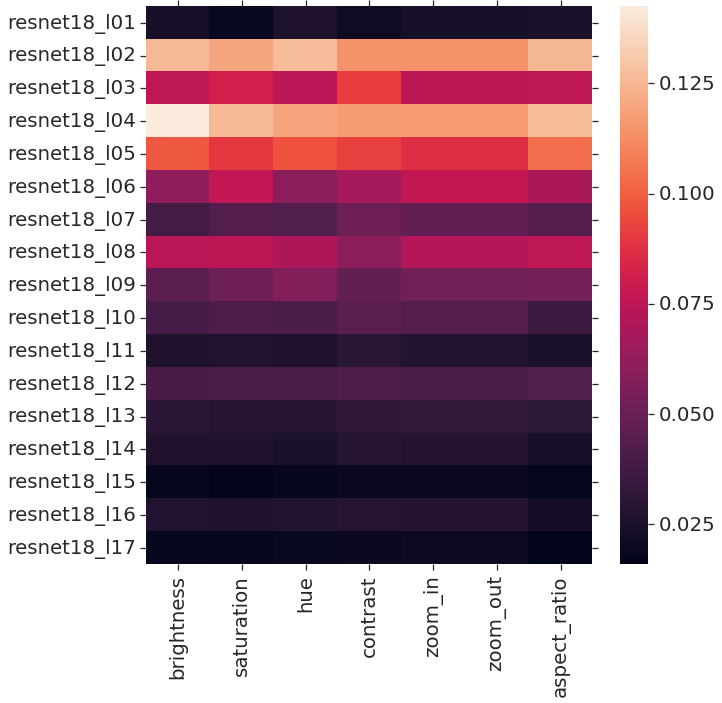}
    \caption{}
    \label{fig:resnet18_mean}
\end{subfigure}
\begin{subfigure}[H]{0.40\textwidth}
    \centering
    \includegraphics[width=\textwidth]{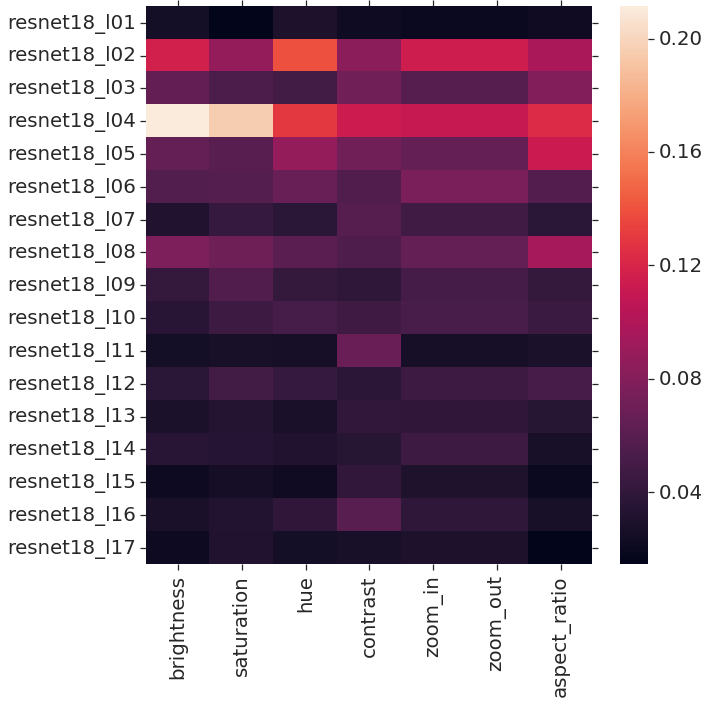}
    \caption{}
    \label{fig:resnet18_max}
\end{subfigure}
\begin{subfigure}[H]{0.40\textwidth}
    \centering
    \includegraphics[width=\textwidth]{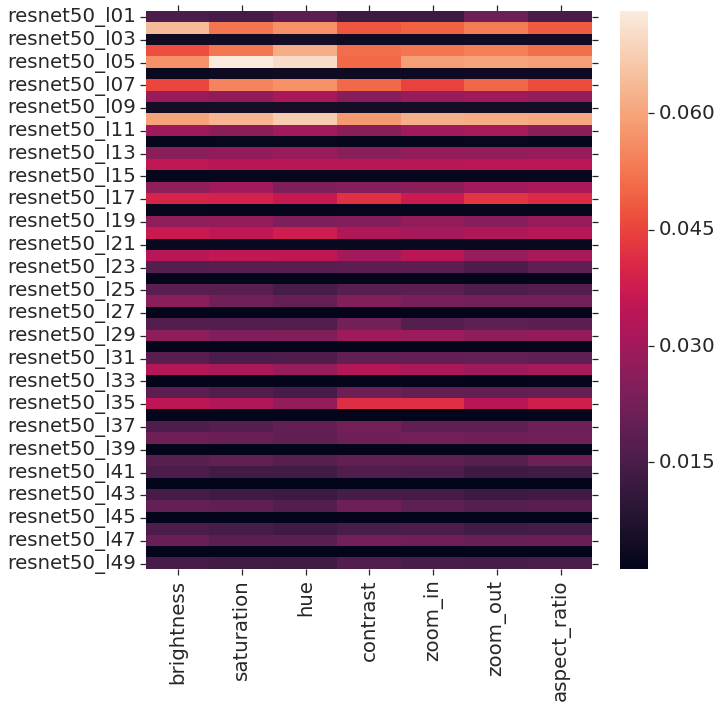}
    \caption{}
    \label{fig:resnet50_mean}
\end{subfigure}
\begin{subfigure}[H]{0.40\textwidth}
    \centering
    \includegraphics[width=\textwidth]{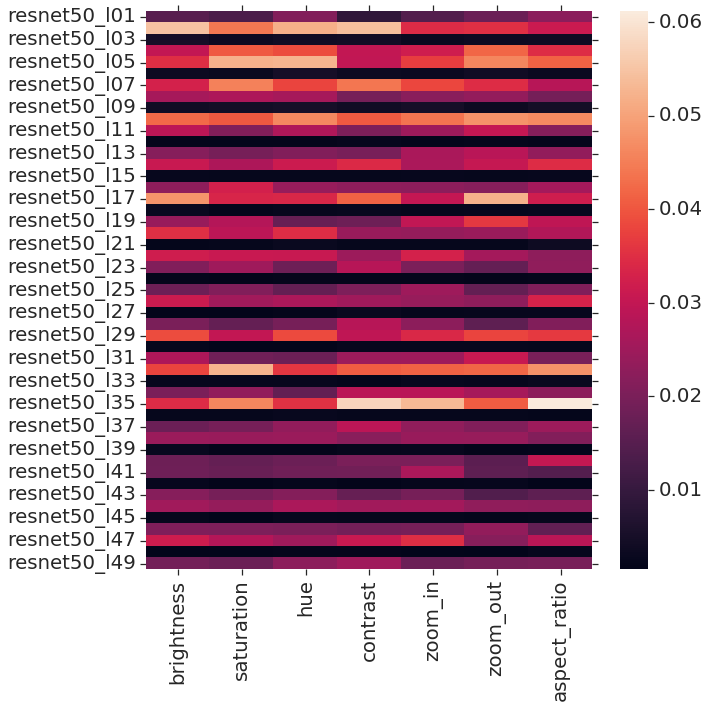}
    \caption{}
    \label{fig:resnet50_max}
\end{subfigure}
\end{center}
\caption{Weightings of activations for ranking tasks with a ResNet-18 backbone (a, b) and ResNet-50 backbone (c, d), with the sum of activations for each task normalized to 1.0. Ranking tasks are ordered from left to right roughly from low-level (color perturbations) to high-level (scale and aspect ratio). Early layers are more important for lower-level ranking tasks, such as color attributes. Color represents mean (a, c) and max (b, d) value across the channel dimension.}
\label{fig:resnet18and50}
\end{figure*}
\autoref{fig:resnet18_mean} and~\autoref{fig:resnet18_max} show the relative importance of ResNet-18 layers for the ranking tasks when taking the mean and max across the channels respectively.
A general trend is that the earlier layers are weighted more highly for all of the ranking tasks.
Interestingly, this trend occurs even when taking the max across channels despite the later layers having more channels than the early layers.

Another difference is that slightly deeper layers appear more important (or alternatively, early layers are less important) for contrast, aspect ratio, and scale (zoom in and zoom out).
This pattern may be the result of contrast, scale, and aspect ratio being a higher-level attribute than brightness and saturation.
While we did not explicitly attempt to build a classification model using backbone features, these differences suggest that backbone features could also be used to classify the the different types of augmentations.

We see a similar trend for the mean (\autoref{fig:resnet50_mean}) and max (\autoref{fig:resnet50_max}) of feature importance across channels for ResNet-50.
For the aspect ratio and zoom in tasks, the most highly weighted layer (when taking the max across channels) occurs later in the model.
In both ResNet-18 and 50, shortcut layers seem to be neglected by the ranking models.
In ResNet-50, the later layers appear to be more highly utilized (especially when taking the maximum across channels) though this effect might be accounted for by ResNet-50's greater number of channels increasing the chances that some channel in a layer may be weighted highly.

To further validate the trend of early layers more strongly encoding augmentation attributes, we rerun a selection of experiments, using activations from only a few layers at a time.
If the early layers are more relevant for encoding augmentation attributes, then we should observe an accuracy drop when using activations from later layers.
Indeed, ~\autoref{tab:halfs} shows this drop, suggesting that even if neural networks encode augmentations, this signal begins to be normalized away in later layers, a trend we discuss further in~\autoref{sec:specialization}.

\begin{table}[tb]
\begin{footnotesize}
\begin{center}
\begin{tabularx}{0.5\textwidth}{  | >{\raggedright\arraybackslash}X 
  | >{\centering\arraybackslash}X 
  | >{\raggedleft\arraybackslash}X |} 
 \hline
 & Zoom In-Train & Zoom In-Val\\ 
  \hline
 ResNet-18 Block 1& 95.5 & 92.2\\
 \hline
 ResNet-18 Block 2& 95.8 & \textbf{92.8}\\
 \hline
 ResNet-18 Block 3& 93.7 & 89.1\\
 \hline
 ResNet-18 Block 4& 93.2 & 90.0\\
 \hline
 ResNet-18 Block 5& 90.0 & 85.1\\
 \hline
 ResNet-18 Block 6& 87.5 & 82.9\\
 \hline
\end{tabularx}
\begin{tabularx}{0.5\textwidth}{  | >{\raggedright\arraybackslash}X 
  | >{\centering\arraybackslash}X 
  | >{\raggedleft\arraybackslash}X |} 
 \hline
 & Aspect Ratio-Train & Aspect Ratio-Val\\ 
  \hline
 ResNet-18 Block 1& 75.7 & 78.7 \\
 \hline
 ResNet-18 Block 2& 87.6 & 86.3 \\
 \hline
 ResNet-18 Block 3& 86.5 & \textbf{88.8}\\
 \hline
 ResNet-18 Block 4& 87.7 & 87.2 \\
 \hline
 ResNet-18 Block 5& 80.1 & 79.1 \\
 \hline
 ResNet-18 Block 6& 66.7 & 62.8 \\
 \hline
\end{tabularx}
\begin{tabularx}{0.5\textwidth}{  | >{\raggedright\arraybackslash}X 
  | >{\centering\arraybackslash}X 
  | >{\raggedleft\arraybackslash}X |} 
 \hline
 & Hue-Train & Hue-Val\\ 
 \hline
 ResNet-18 Block 1& 75.1 & \textbf{77.5} \\
 \hline
 ResNet-18 Block 2& 77.6 & 76.6 \\
 \hline
 ResNet-18 Block 3& 77.8 & 73.0 \\
 \hline
 ResNet-18 Block 4& 81.0 & 72.5 \\
 \hline
 ResNet-18 Block 5& 82.5 & 67.5 \\
 \hline
 ResNet-18 Block 6& 82.1 & 65.8 \\
 \hline
\end{tabularx}
\caption{Ranking accuracy when only using features from a single block of ResNet-18, ordered from early to later layers. The early blocks yield higher accuracy, indicative of early layers more strongly encoding augmentation attributes.}
\label{tab:halfs}
\end{center}
\end{footnotesize}
\end{table}

\section{Discussion}

\paragraph{Specialization vs. normalization}
\label{sec:specialization}
For augmentations that are encoded or captured by CNN activations, we ask \emph{where} or at what depth?
We describe this question as the specialization vs. normalization question: we posit that data augmentations that are encoded by earlier layers are \emph{normalized} away by the model, whereas attributes that are encoded in later layers incur \emph{specialization}.
Intuitively, if a model captures augmentation attributes in early layers but discards this information by the later layers, it has normalized away the augmentation.
However, if a model retains augmentation differences in later layers, the intuition is that this augmentation incurs specialization in the same way that the last layer is specialized at a per-class granularity for classification tasks.


The importance of activations from earlier layers relative to those from later layers for our ranking objectives suggests that attributes such as scale are normalized away by CNNs.
This phenomenon appears more desirable than the alternative where augmentation attributes are encoded and preserved throughout the model, indicating limited generalization at the output.
The lower ranking accuracy when using a ResNet-50 backbone (vs. ResNet-18) may indicate that more accurate models do a better job of normalizing away augmentations.

\paragraph{An adversarial ``ranking model''}
An alternative scenario we considered was a generative model that proposes augmented images that attempt to fool the backbone model, taking activations of a pre-trained backbone as input.
However, a difficulty of this approach is that some popular augmentations (scale transformations) are not easily expressible using standard vision operators or are not differentiable.
Still, we see adversarial augmentations as an important related problem: what augmentations are the most difficult for current models?

\paragraph{Can ranking objectives be used as pre-training tasks?}
\begin{figure}
    \centering
    \includegraphics[width=0.40\textwidth]{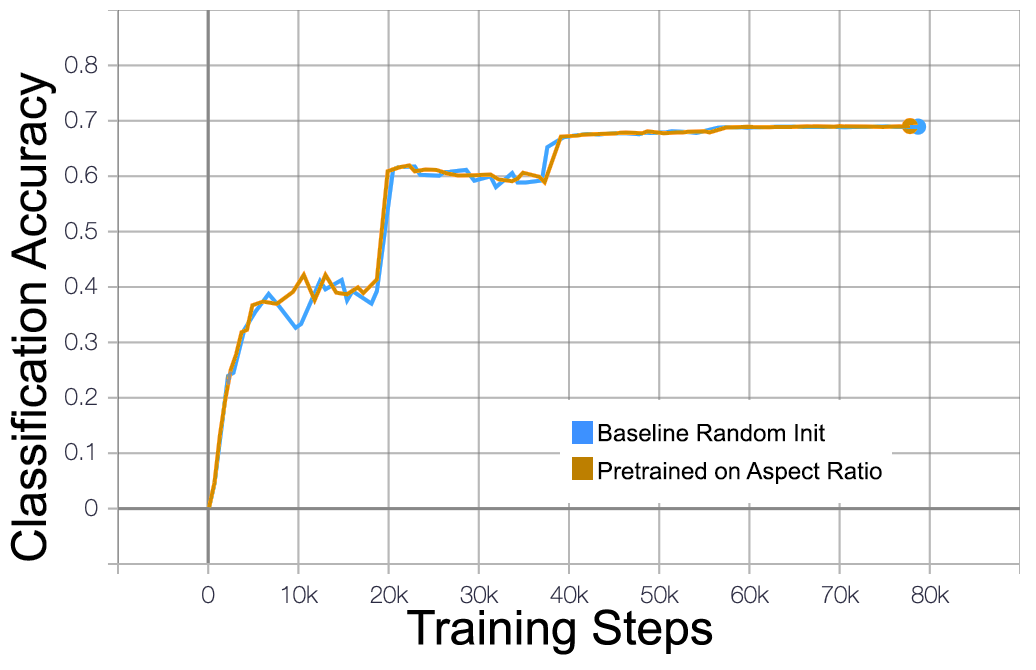}
    \caption{ImageNet classification accuracy vs. training steps of a from-scratch model compared to that of a backbone model pre-trained on the aspect ratio ranking task. Classification performance does not improve, suggesting that encoding augmentations is not inherently desirable.}
    \label{fig:pretraining}
\end{figure}
That neural networks appear to encode data augmentation transformation attributes raises the question of whether these attributes are inherently useful for vision tasks.  
If it is useful for neural network models to encode these attributes, would a source of accurate scale, aspect ratio, or color information improve their performance?
~\autoref{fig:pretraining} shows the results of an experiment where a backbone is pre-trained without class labels via a downstream ranking task (aspect ratio).
We find that pre-training to rank augmentations does not improve classification performance (with no improvement over training from scratch).
The lack of improvement seems to support the hypothesis that encoding augmentations is not inherently desirable, and that normalization is the desired effect.


\paragraph{Limitations and future work}
In using a simple linear layer to build our ranking model, we sacrifice model performance for interpretability.
It may be possible that with sufficient representation power (e.g., with a deeper or more complex architecture) in the ranking model, data augmentation transformations can be recovered with high accuracy using only deep network layers.
Still, we believe that using a linear ranking model reveals that augmentation transformations are prominent in neural network features in early layers.
A natural extension of this work would include novel model architectures and augmentations.
\section{Conclusion}
We posed the question of whether modern CNNs encode attributes corresponding to popular data augmentations such as color and scale transformations.
To answer this question, we proposed data augmentation ranking tasks to understand whether CNNs encode differences introduced by data augmentations and designed a method that compares the predictive power of the intermediate activations in a CNN.
We find that CNNs encode many data augmentations, and that the earlier layers are generally the most predictive of augmentation transformations.
Our findings also suggest that the signal of augmentations fades in later layers, and that more accurate models normalize away augmentations to a greater extent.

{\small
\bibliographystyle{IEEEtran}
\bibliography{main}
}
\end{document}